\documentclass[pdflatex,sn-mathphys]{sn-jnl}

\jyear{2022}%
\usepackage[utf8]{inputenc} 
\usepackage[T1]{fontenc} 
\usepackage[english]{babel}

\theoremstyle{thmstyleone}%
\theoremstyle{thmstyletwo}%

\theoremstyle{thmstylethree}%

\usepackage{caption}
\usepackage{subcaption}

\usepackage[inline]{trackchanges}
\usepackage{hyperref}

\begin{document}

\title[LiDAR-based drone navigation with reinforcement learning]{LiDAR-based drone navigation with reinforcement learning}


\author{\fnm{Paweł} \sur{Miera}} \email{miera@student.agh.edu.pl}
\author{\fnm{Hubert} \sur{Szolc} \href{https://orcid.org/0000-0003-3018-5731}{\includegraphics[width=10pt]{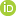}}} \email{szolc@agh.edu.pl}
\author*{\fnm{Tomasz} \sur{Kryjak}* \href{https://orcid.org/0000-0001-6798-4444}{\includegraphics[width=10pt]{orcid.png}}} \email{tomasz.kryjak@agh.edu.pl}

\affil{\orgdiv{Embedded Vision Systems Group, Computer Vision Laboratory, Department of Automatic Control and Robotics}, \orgname{AGH University of Science and Technology}, \orgaddress{\street{al. Mickiewicza 30}, \city{Krakow}, \postcode{30-059}, \country{Poland}}}

\abstract
{Reinforcement learning is of increasing importance in the field of robot control and simulation plays a~key role in this process.
In the unmanned aerial vehicles (UAVs, drones), there is also an increase in the number of published scientific papers involving this approach.
In this work, an autonomous drone control system was prepared to fly forward (according to its coordinates system) and pass the trees encountered in the forest based on the data from a~rotating LiDAR sensor.
The Proximal Policy Optimization (PPO) algorithm, an example of reinforcement learning (RL), was used to prepare it.
A~custom simulator in the Python language was developed for this purpose.
The Gazebo environment, integrated with the Robot Operating System (ROS), was also used to test the resulting control algorithm.
Finally, the prepared solution was implemented in the Nvidia Jetson Nano eGPU and verified in the real tests scenarios.
During them, the drone successfully completed the set task and was able to repeatably avoid trees and fly through the forest.
}

\kewwords{LiDAR, Reinforcement learning, RL, Drones, Automatic control, ROS, Gazebo}



\maketitle

\section{Introduction}\label{sec:wprowadzenie}

Unmanned aerial vehicles (UAVs), also known as drones, can most often be seen in the form of aircrafts or multi-rotors.
They are characterised by an extremely wide range of potential applications:
\begin{itemize}
    \item inspection of inaccessible areas  \cite{MANDIROLA2022102824},
    \item transport of small consignments, for example rapid delivery of blood or delivery of medical supplies to hard-to-reach areas \cite{8701196},
    \item monitoring and analysis of terrain based on an orthophoto created from images from a~camera placed on a~drone \cite{CARABASSA2020110717},
    \item monitoring of traffic congestion in the city \cite{8730677}.
\end{itemize}

Most of these cases make at least some use of an automatic control system.
In the simplest view, it involves flying sequentially to preset points whose coordinates consist of: longitude and latitude, angle of rotation and altitude. 
Carrying out such a~mission requires the flight controller to have continuous access to the GPS (Global Positioning System) signal, through which it is possible to obtain the vehicle's global position using information provided by satellites orbiting the Earth.
In general, however, the drone can perform a~mission in various types of enclosed spaces, such as a~cave \cite{DBLP}, a~tunnel as well as a~warehouse or hall.
Then it does not have access to the aforementioned GPS signal.
In such places, suitable sensors can be used to determine the position. 
These allow to obtain the current position relative to the starting point, which is calculated from the fusion of data from the inertial and vision sensor.

Currently, many research papers consider drone control by using reinforcement learning.
This approach involves training a~so-called 'agent' to perform a~specific task. 
The training (usually simulation-based) aims to maximise the return received for performing actions in the environment. 
The trained agent's policy can then be transferred to the real world and be responsible, for example, for controlling the drone's speeds in the X, Y and Z axes, also in the absence of a~GPS signal.

In this paper, we present the use of reinforcement learning for the task of navigating a~drone based on data returned by a~rotating LiDAR sensor.
For this purpose, we applied the Proximal Policy Optimization (PPO) algorithm.
The test case is to fly in a~straight line (in a~given reference system) through a~forest, so that there are no collisions with trees.
We trained the agent in a~custom-built simple simulator, then implemented it in Nvidia's Jetson Nano chip and conducted real-world tests.

The results of the conducted research should be considered satisfactory.
Our proposed method makes it possible to obtain an effective drone control algorithm in a~forest environment characterised by a~significant accumulation of tree obstacles.
Moreover, it is relatively cheap and accessible, as it does not require the use of complex simulation environments or advanced algorithms for the fusion of data from various sensors.

The rest of the paper is organised as follows.
Section \ref{sec:literatura} provides an overview of the scientific literature related to the addressed subject.
We paid particular attention to works that used reinforcement learning.
Section \ref{sec:metoda} includes a~description of our proposed drone control method.
We include there details of the used simulation environment, the developed agents and the specification of the hardware system we used for the real-world tests.
Section \ref{sec:rezultaty} presents the results obtained with the proposed control method.
It describes both the outcome of training in simulation and the effects of transferring the system to reality.
The final Section \ref{sec:podsumowanie} summarises the paper and indicates plans for further development of the work.

\section{Related work}\label{sec:literatura}
The use of reinforcement learning for the task of controlling different robots is currently a~very popular approach.
This is reflected in the relatively large number of new research papers on the subject.

The article \cite{slm_lab} presents a~comparison of the results of different reinforcement learning algorithms in various environments. 
They were divided into those involving discrete control (the action vector can only have certain values) and continuous control (the values of the action vector can be arbitrary). 
In both cases, the PPO algorithm performed very well, second only to SAC (Soft Actor Critic) in continuous environments, which, however, needed longer training times to achieve good results.

The paper \cite{9537063} describes the application of reinforcement learning to the task of safe lane changing by a~car in a~multi-vehicle traffic situation on the road. 
For this purpose, a~simulation was prepared in the Unity tool and the PPO and SAC algorithms were trained. 
In the final comparison, the effectiveness of both algorithms was more than 90\%, but the training time of the former was much shorter.

Reinforcement learning also has increasing number of applications in the field of drones, as described in the review article \cite{przeglad}.
Its authors show a~growing interest in using these algorithms to control the speed and altitude of a~multi-rotor aircraft.
However, they also highlight the problems associated with transferring such systems from simulation to reality. 
These include hardware limitations and the differences between the two environments (sim-to-real gap).

The paper \cite{ladowanie} presents the development of a~system whose mission was to land a~drone on a~moving platform. 
This task was accomplished by using an RL agent for which the observations were the current states of the vehicle. 
After training in the simulation, it was launched at the ground station and control signals were transmitted via Wi-Fi network. 
The effectiveness of the operation was confirmed both in the simulation environment and in reality.

Another example of the use of reinforcement learning for UAV control is the work presented in paper \cite{racing}, in which the authors demonstrated how to determine the trajectory of a~drone's flight through gates in a~changing environment. 
This task was performed by training an agent with the PPO algorithm in the Fligthmare simulator. 
The policy was then tested on one thousand flights, and the final failure rate for the best network was 0.6\%.
One of the generated trajectories was also run on an actual drone, but there were large errors in position tracking during the flight.

Analysis of the above-mentioned articles showed that the PPO algorithm has good performance and acceptable learning time.
Therefore, we decided to use it in our research.

\section{Proposed method}\label{sec:metoda}
This paper focuses on the use of reinforcement learning for the task of controlling a~drone.
The considered problem concerns the flight through a~previously unknown spatial environment in the form of a~forest.
The proposed approach can be divided into three steps:
\begin{enumerate}
    \item Preparation of the simulation environment.
    \item Development of the agents performing the stated task.
    \item Implementation of the obtained control algorithm in the hardware system.
\end{enumerate}

We provide details of the realisation of each of these later in this section.
For the implementation of the system, we assumed that the drone receives information about its surroundings using the RPLIDAR A2 sensor \cite{RPLIDAR}.
This is a~device that uses a~laser sensor on a~rotating base to measure distance. 
Its result is a~contour scan of the environment, consisting of distances to the nearest objects for specific angles of rotation.
The range is 16 metres, while the angular resolution is 0.225\textdegree{}.

\subsection{Simulation environment}\label{subsec:symulator}

In order to train the agents in a~best possible way, we decided to use two simulators.
The first one, prepared from scratch in Python, was intended to support the learning process.
The second simulator was used to make the adjustments to the algorithm needed to transfer the system to a~hardware chip.
For this, we used the off-the-shelf Gazebo Simulator environment.

We started the implementation of the first simulator by creating a~drone class as an object that moves in a~changing manner. 
We present the method for calculating acceleration in Equation \eqref{eq:drone_acc} -- we restrict its value to the range $[-max\_acc, max\_acc]$. 
We calculate the current speed and position of the drone from Equations \eqref{eq:drone_speed} and \eqref{eq:drone_pos}, respectively (similarly for each axis of the rectangular coordinate system).

\begin{equation}
    \label{eq:drone_acc}
    a = \frac{v_{req} - v_0}{dt}
\end{equation}

\begin{equation}
    \label{eq:drone_speed}
    v = v_0 + a * dt
\end{equation}

\begin{equation}
    \label{eq:drone_pos}
    x = x_0 + v * dt + \frac{a * dt^2}{2}
\end{equation}
where: $a$ -- acceleration [m/s$^{2}$], $v_{req}$ -- target velocity [m/s], $v_0$ -- velocity in the previous simulation step [m/s], $dt$ -- time of one simulation step [s], $v$ -- actual velocity [m/s], $x$ -- actual position [m], $x_0$ -- position in the previous simulations step [m].

\begin{figure}[t]
	\centering
    \hspace{-30pt}
	\begin{subfigure}{0.45\textwidth}
		\includegraphics[width=66mm]{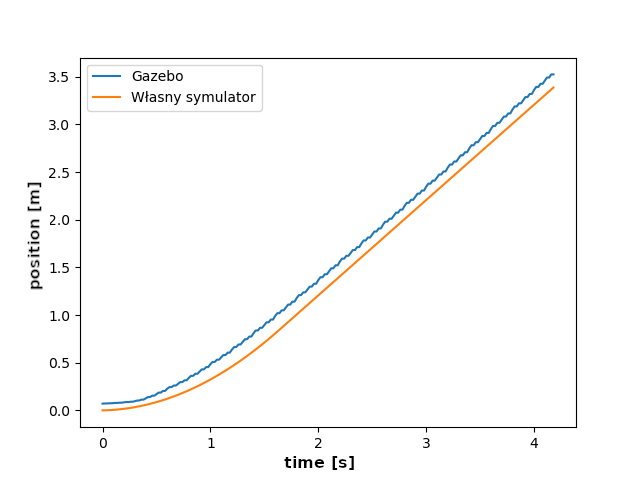}
		\subcaption{}
		\label{img:p_my_gaz}
	\end{subfigure}
    \hspace{22pt}
	\begin{subfigure}{0.45\textwidth}
		\includegraphics[width=66mm]{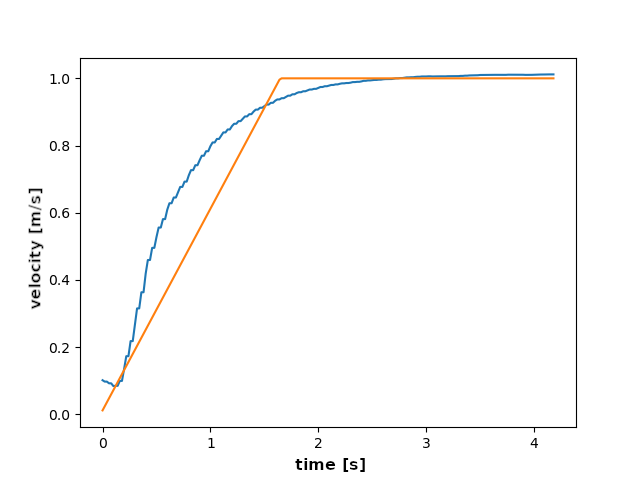}
		\subcaption{}
		\label{img:v_my_gaz}
	\end{subfigure}
	
	\caption{Comparison of the values between our simulator (orange) and Gazebo environment: \protect\subref{img:p_my_gaz} position, \protect\subref{img:v_my_gaz} velocity along the axis Y.}
    \vspace{-5pt}
\end{figure}

We selected the coefficients in Equations \eqref{eq:drone_speed} and \eqref{eq:drone_pos} so that, for the same enforcing, the change in position and velocity in the developed drone simulator is as similar as possible to the values obtained from Gazebo. 
An example of the results obtained in this way on the Y-axis is shown on the graph in Figures \ref{img:p_my_gaz} and \ref{img:v_my_gaz}.
Despite the noticeable differences, the obtained accuracy proved sufficient for further experiments.

\begin{figure}[t]
	\centering
    \includegraphics[width=0.49\textwidth]{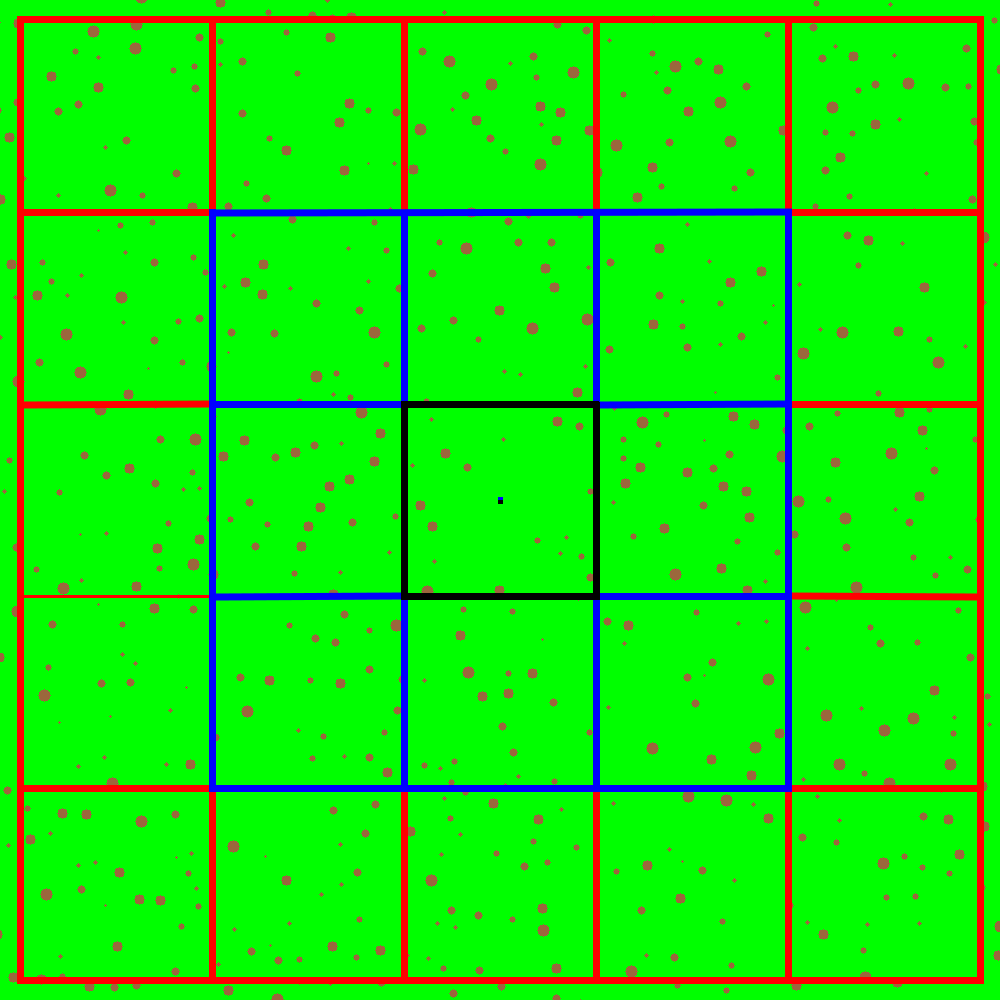}
    \vspace{5pt}
	\caption{Map of the forest simulation with the grid plotted. The black colour identifies the cell in which the drone is currently located (centre of figure), the blue colour -- cells considered in the current simulator step, and the red colour -- cells ignored to speed up the calculation.}  
	\label{img:grid}
    \vspace{-5pt}
\end{figure}

An important functionality of our simulator is the ability to generate a~virtual forest.
In Figure \ref{img:grid}, we show a~random map with an applied grid with a~side length of 16 metres. 
The trees within it were assigned to each cell.
Their attributes are the coordinates of the centre position and the radius of the branch.
We applied the division into separate sectors so that, when searching for trees within the range of the laser, only those belonging to the current drone cell (black) and neighbouring cells (blue) are taken into account. 
The rest of the grid (in red) is not taken into account in a~given calculation step, thus speeding up the computation.

Simulation launch starts with the initialisation of variables and the creation of a~drone object with zero values for position and velocity. 
Trees are then randomly generated in each grid cell, taking into account the following parameters:
\begin{itemize}
    \item \texttt{trees\_per\_grid} -- specifies the number of trees that can be located in one grid cell (a~value between 30 and 60, drawn after each iteration);
    \item \texttt{tree\_radius\_range} -- defines the range of tree radii in metres;
    \item \texttt{trees\_min\_distance} -- determines the minimum distance between the generated trees.
\end{itemize}
We selected the values of the coefficients in such a~way that the simulation resembled the real forest as much as possible.
If the software could not find a~new location for the next obstacle for 1000 trials, the loop was interrupted, resulting in fewer trees in the current grid.

We also prepared a~simulation of the RPLIDAR device.
We obtained the resulting measurements by finding the common points of the equations of the straight line and the circle.
For this purpose, we used the fact that the straight line passes through the point $P_0$, whose coordinates were equal (in value) to the difference between the position of the drone and the centre of the tree.
The developed Python implementation of the aforementioned equations also rejects unwanted solutions resulting from the intersection of the circle at two points and the same tangent value for angles shifted by multiples of $P_0$.

\begin{figure}[!t]
	\centering
    \includegraphics[width=0.49\textwidth]{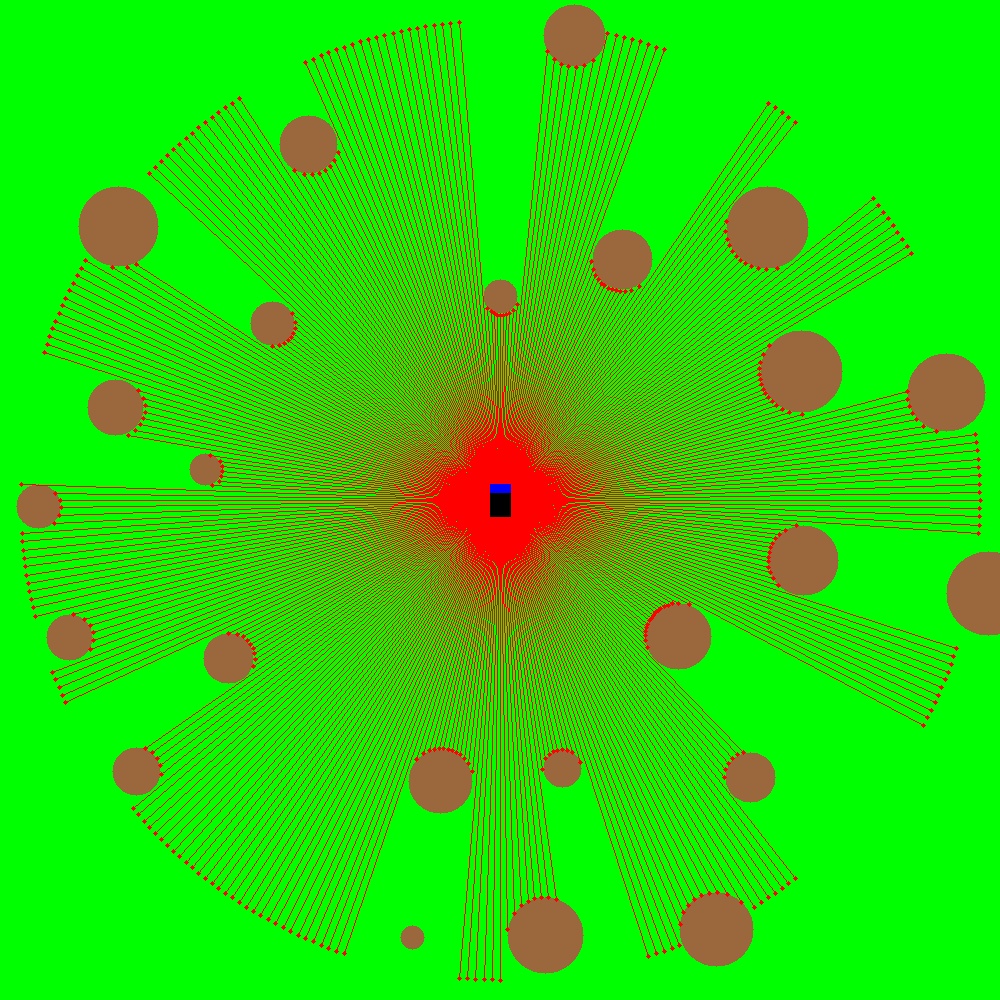}
    \vspace{5pt}
	\caption{Display window of the prepared simulation.}  
	\label{img:drone_game}
    \vspace{-5pt}
\end{figure}

The final design of the simulation window is shown in Figure \ref{img:drone_game}.
We used the prepared simulator to train RL agents (Section \ref{subsec:RL}).
We also made its source code available in the Github repository\footnote{\url{https://github.com/vision-agh/python_drone_game_laser_scan}}.

We also used the Gazebo environment when transferring the algorithms to the hardware platform.
The drone is controlled there by communicating with the SITL (software-in-the-loop) mode of the PX4 software.
This allows the code prepared in the simulation to then run unchanged on the actual drone. 
The whole system is also integrated into the ROS (Robot Operating System), which facilitates communication between all the added sensors.

\begin{figure}[!t]
	\centering
    \includegraphics[width=0.49\textwidth]{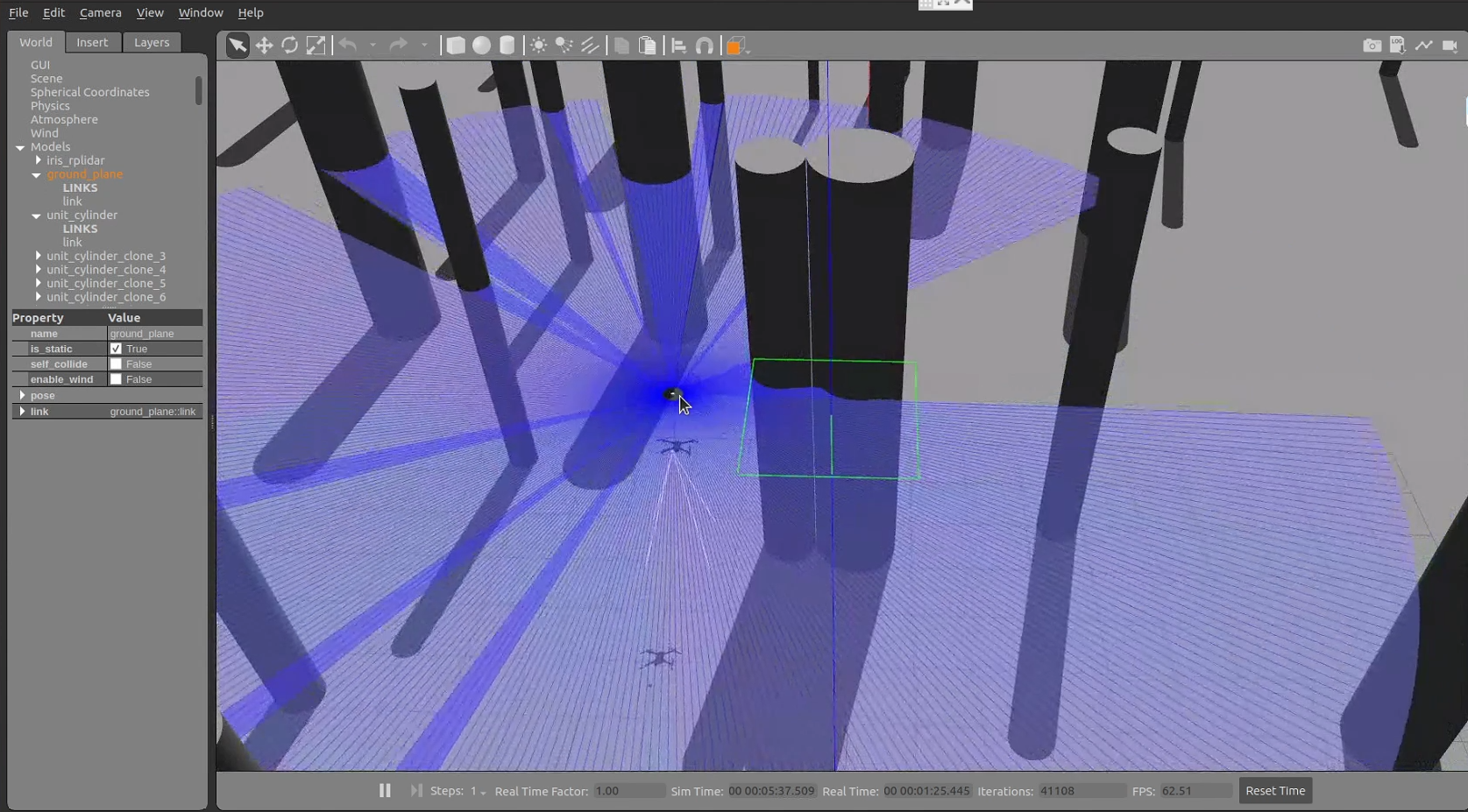}
    \vspace{5pt}
	\caption{A~map we prepared to simulate a~drone with the RPLIDAR sensor in the Gazebo environment.}  
	\label{img:gazebo_lidar}
    \vspace{-5pt}
\end{figure}

In order to use the Gazebo simulator in the discussed problem, we prepared a~special map. 
The trees are generated on it as tall cylinders with different radii.
For the drone and the RPLIDAR sensor, we used the ready-to-use models available with the PX4 software.
However, we changed some parameters in the default configuration file, such as resolution, sampling frequency, height relative to the base and laser range.
The prepared simulation window is shown in Figure \ref{img:gazebo_lidar}.

\subsection{Developed RL agent}
\label{subsec:RL}

We used reinforcement learning algorithms to control the drone. 
In general, this approach involves a~so-called agent performing some action in the environment based on the observations it receives and the value of the reward.
At the same time, it changes the state of the environment in this way, which influences its subsequent decisions.
The agent takes these decisions by means of a~policy, i.e. a~function that returns a~certain action based on the observations. 
It is often implemented as a~neural network whose parameters are changed during training. 
Its architecture may consist of several layers that are fully connected, but in some environments recursive layers are also applicable (they have memory, so that previous calculations affect the current result). 
When the observation is an image, the neural network model may also include convolutional and pooling layers.

The simplest algorithm for reinforcement learning is Q-learning.
The selection of the next action to be performed in it is done using a~so-called Q-Table.
This contains the return (discounted sum of rewards) that will potentially be received after performing a~particular action in a~particular state and making further decisions according to the used policy.
This is known as the action-value function.
The values of the Q-Table cells are updated iteratively using the Bellman equation \cite{bellman}. 
Q-learning interacts with the environment using a~different policy to the one it is optimising. 
A~certain modification of this approach is the SARSA (State-Action-Reward-State-Action) algorithm.
One and the same policy is both updated and used in it to perform an action.
Popular modern reinforcement learning algorithms have been developed on the basis of the discussed approaches:
\begin{itemize}
    \item Deep Q~Network (DQN),
    \item Proximal Policy Optimization (PPO),
    \item Soft Actor Critic (SAC),
    \item Trust Region Policy Optimization (TPRO),
    \item Advantage Actor Critic (A2C).
\end{itemize}

Based on a~literature review (Section \ref{sec:literatura}), in this work we decided to use the PPO algorithm. 
To begin with, we assumed that the goal of the drone was to fly 30 metres through the forest along the X-axis (in a~given coordinate system), without colliding with any tree. 
The PPO agent received only readings from the rotating LiDAR sensor.
Therefore, the observations did not include any data on the drone's current speed or position. 
They were processed by the agent's policy, which we implemented through a~neural network.
In order to fit the observations into the range $<-1; 1>$ that it accepts, we decided to normalise them according to Equation \eqref{eq:normalisation}:
\begin{equation}
    \label{eq:normalisation}
    r = (r - h_2) / h_2
\end{equation}
where: $r$ -- distances from the rotating LiDAR [m], $h_2$ -- half of the maximum LiDAR range [m].

We also developed a~cost function that was the sum of the following components:
\begin{itemize}
    \item penalty for approaching a~tree at a~distance of less than 0.15 m --- value: $-0.25$;
    \item penalty for colliding with a~tree --- value: $-1.5$;
    \item penalty for moving away from the centre-line of the environment (relative to the Y-axis), the value of which was $-0.1 * \|p_y\|$, where $p_y$ is actual drone's Y~coordinate;
    \item reward for the high speed in the X~axis ($v_x$), calculated as $0.8 * v_x$ --- in case the drone was flying in the wrong direction, its value was multiplied by $-3$.
\end{itemize}
In the role of both policy and value function, we used a~multilayer perceptron (MLP).
In both cases, it consisted of two hidden layers with ReLU (rectified linear unit) activation.
Their sizes were 128 neurons for the policy and 256 for the value function.

We implemented the training of the PPO agent through a~script written in Python. 
We first initialised the prepared simulator (Section \ref{subsec:symulator}) and the RL agent model with it, and then ran the training for a~set number of episodes.

\subsection{Hardware system}

\begin{figure}[t]
	\centering
    \includegraphics[width=0.49\textwidth]{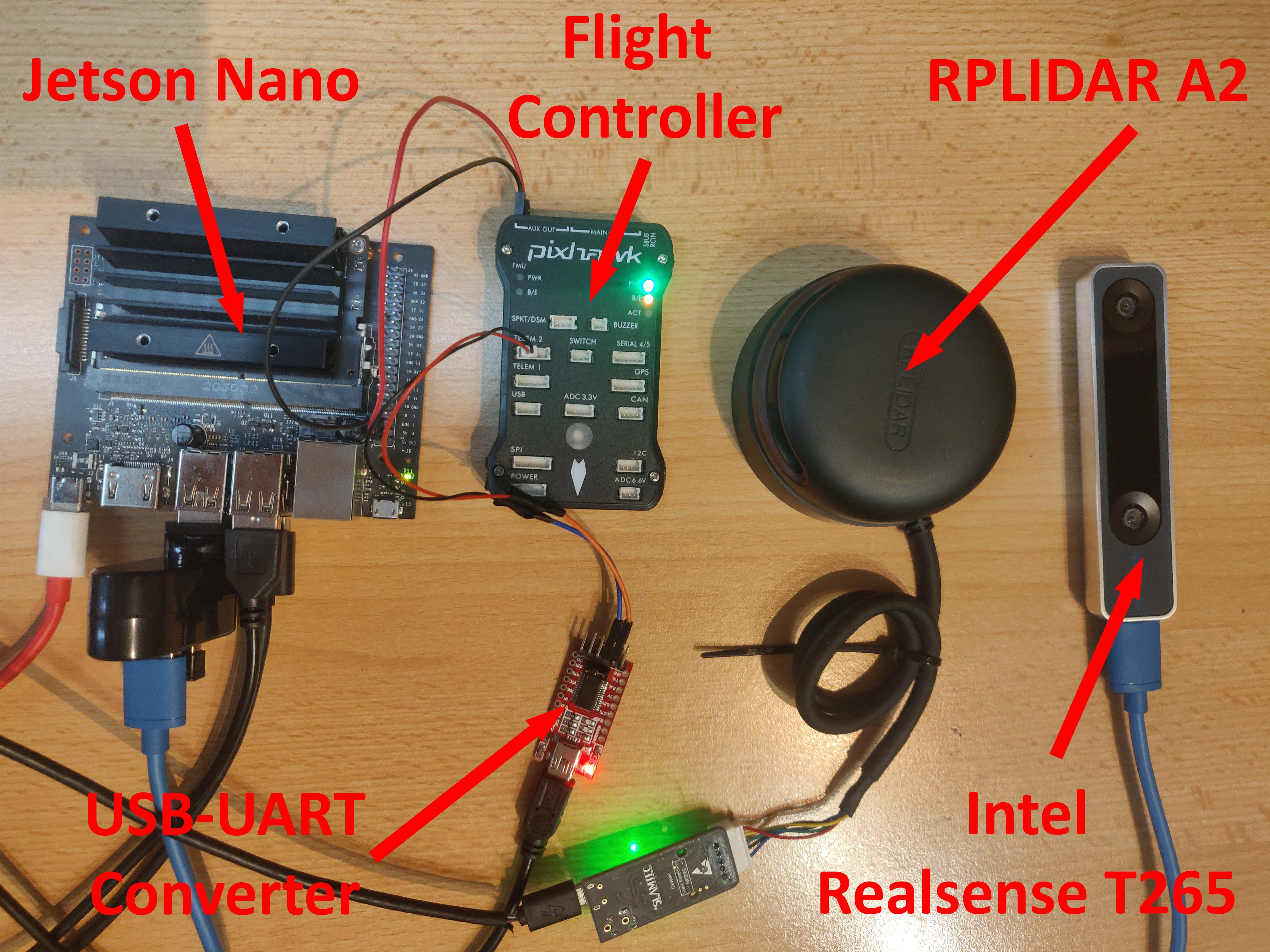}
    \vspace{5pt}
	\caption{The hardware components we used to control the drone.}  
	\label{img:components}
    \vspace{-5pt}
\end{figure}

After the training process, we implemented the final RL agent in a~Jetson Nano hardware chip, which acted as the flight computer.
It communicated with the Black Cube flight controller, to which we uploaded the PX4 software.
Data exchange between these devices took place via a~serial port using a~USB-UART converter.
In addition, we used the aforementioned RPLIDAR A2, as well as the Intel RealSense T265 camera.
The latter device was used to provide information on the current location of the vehicle required by the Black Cube flight controller.
All the components communicating with the on-board computer are shown in Figure \ref{img:components}.

During the implementation of the control algorithms, we used the Ubuntu 18.04 operating systems and ROS Noetic (which we ran in a~specially prepared Docker container).
This ensured a~high degree of similarity between the development environment on the target hardware platform and the Gazebo simulator used and contributed to facilitating the entire migration process.

\section{Results}\label{sec:rezultaty}

Our method of using reinforcement learning for the drone control task (Section \ref{sec:metoda}) implies a~two-stage testing process.
First, it takes place at the simulation phase, during the training of the agent.
Then, the prepared system is verified in a~real operational environment.
The results of the two stages mentioned are presented below.

\subsection{Agent's training}

In executing the training process, we used the implementation of the PPO algorithm in the library \textit{stable-baselines3} \cite{stable-baselines3}.
We ran the calculations on PCs, each equipped with an Nvidia GeForce RTX 3060 GPU.
The trained RL agents were subjected to the evaluation consisting of 100 flight missions through a~forest generated using a~simulator that we prepared (Section \ref{subsec:symulator}).
Each time, the placement and size of the trees were drawn again to test the control algorithm on as many cases as possible.
We considered a~flight successful when the drone flew an assumed distance of 30~m, avoiding collisions with other objects along the way.
The highest achieved success rate (according to the presented criterion) was \textbf{91\%}.

We then also conducted similar experiments using the Gazebo simulator.
In this case, we mainly investigated whether the agent behaves correctly in a~software environment close to the target.
Due to the static nature of the map, we decided to perform 5~test flights.
The experimental results confirmed the correctness of the implementation -- the agent reproducibly avoided collisions with trees.

\subsection{Real-world flights}

\begin{figure}[!t]
	\centering
    \includegraphics[width=0.49\textwidth]{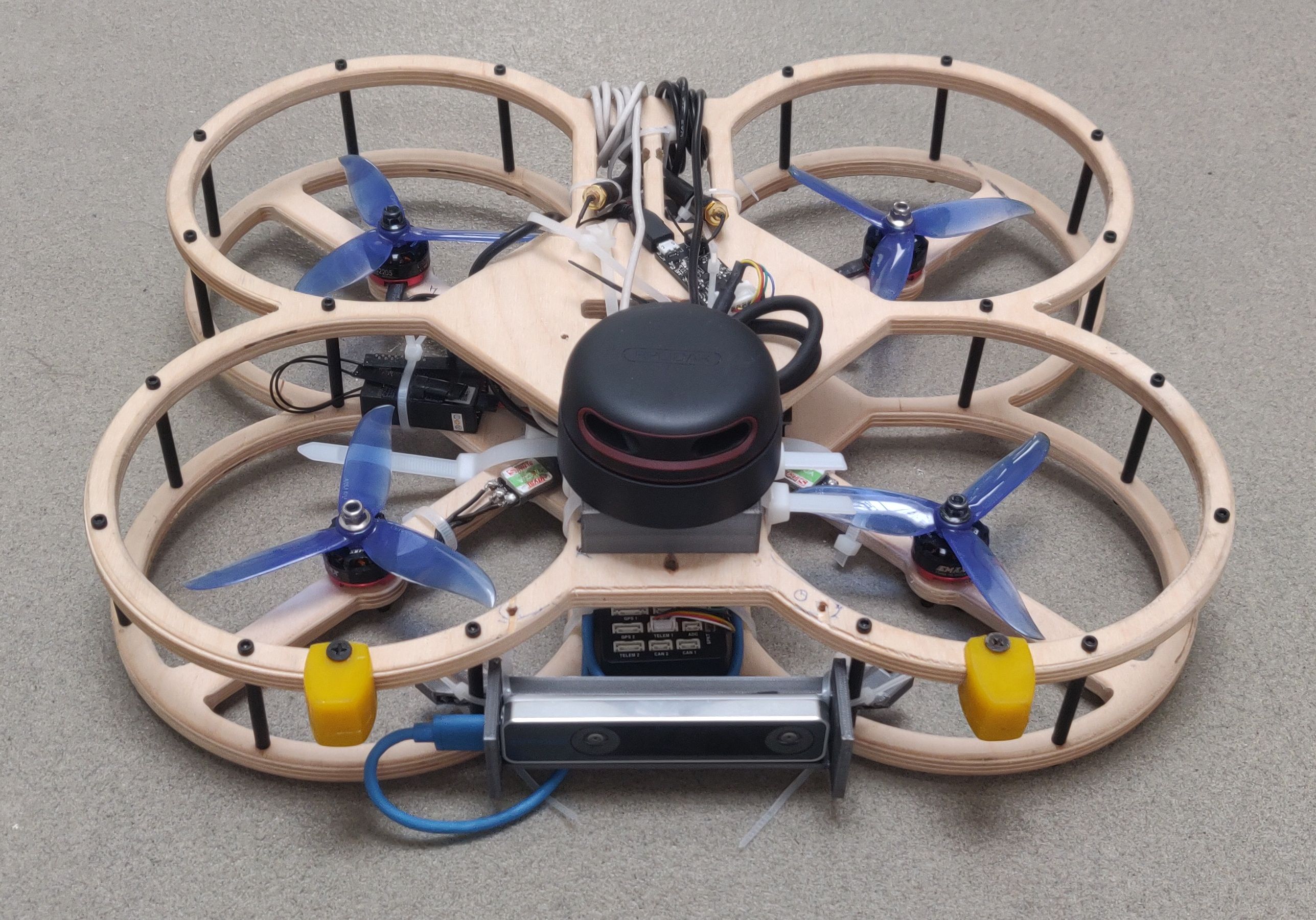}
    \vspace{5pt}
	\caption{Drone's launch configuration for the real-world test scenarios.}  
	\label{img:drone}
    \vspace{-5pt}
\end{figure}

For the real-world flights, we used a~quad-rotor with a~frame made of wood. 
The whole design had a~relatively high weight, resulting in a~low thrust-to-weight ratio and short flight times. 
We present the drone in its launch configuration in Figure \ref{img:drone}.

We conducted the test flights in a~forest, in a~location as similar as possible to the conditions assumed in the simulation. 
The drone was likely to encounter mainly trees with a~small number of low branches on its route, while medium-height shrubs were not very common.
We conducted the missions from different starting points in such a~way that the vehicle was always directed into a~similar area of the forest. 
The goal of the mission was to fly a~preset distance along the X-axis (in a~given coordinate system) without hitting a~single tree. 
We additionally supervised all tests via a~radio controller, which made it possible to take over manually in case of emergency. 

In total, we carried out 25 test flights.
Their results are presented in Table \ref{tab:rezultaty rzeczywistosc}.
The vast majority of the test flights (\textbf{80\%}) were successful.
The drone flew the preset distance through the forest, avoiding most of the trees encountered along the way.
It should be emphasised that its construction wasn't damaged in any case.
\begin{table}[!t]
    \centering
    \caption{Results of the real-world test scenarios.}
    \label{tab:rezultaty rzeczywistosc}
    \begin{tabular}{|l|c|c|} 
    \hline
    Total flights & 25 & 100\% \\
    \hline
    Without hitting a~tree & 13 & 52\% \\
    \hline
    Continued after hitting a~tree & 7 & 28\% \\
    \hline
    Ended after hitting a~tree & 5 & 20\% \\
    \hline
    \end{tabular}
\end{table}

However, the flight efficiency itself is noticeably lower than that previously obtained in the simulator.
This exemplifies the problem referred to in the literature as the \textit{sim-to-real gap}.
Note that in the discussed case, the simulation used simple geometric figures with a~regular shape, which can be considered an extremely simplified tree model.
Despite this, the RL agent was able to build a~sufficiently generic view of the environment on this basis, which enabled multiple successful flights through the forest.
In this context, the obtained result should be considered satisfactory at this stage of the research.

\begin{figure}[!t]
	\centering
    \includegraphics[width=0.49\textwidth]{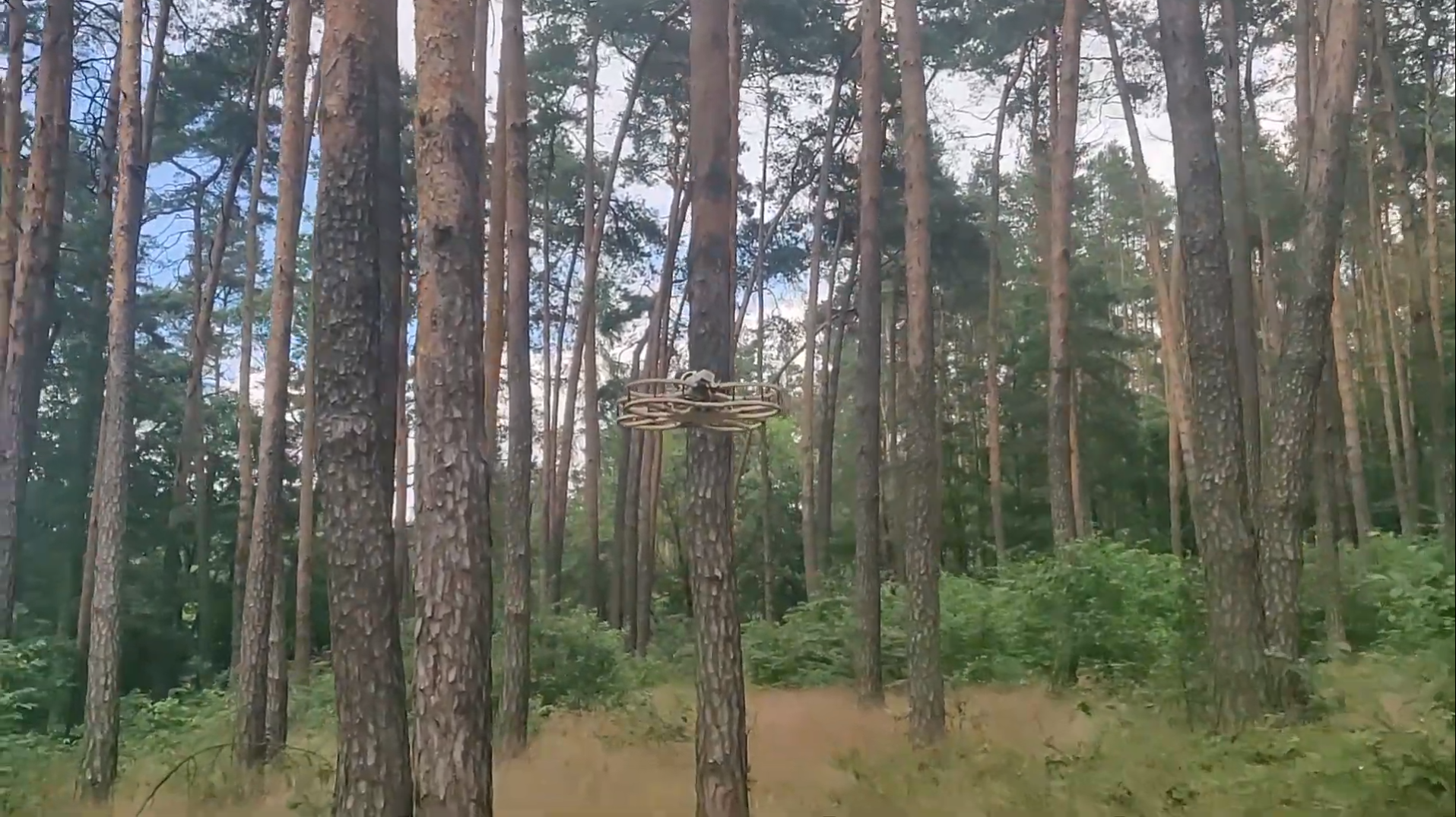}
    \vspace{5pt}
	\caption{A~frame from the video showing one of the real-world test flights.}  
	\label{img:forest}
    \vspace{-5pt}
\end{figure}

Figure \ref{img:forest} shows a~photo of one of the real-world flights.
A~video showing a~summary of the system's operation can be seen on YouTube\footnote{\url{https://www.youtube.com/watch?v=JqosupMgu7g}}.

\section{Summary}\label{sec:podsumowanie}

In this paper, we presented the use of reinforcement learning algorithms to implement a~drone control system that was intended to autonomously fly forward and avoid encountered trees in the forest based on measurements from a~rotating LiDAR sensor. 
When training the RL agent, we used a~PPO algorithm.
To do this, we developed our own simulator written in Python, which mapped the forest environment in a~simplified way.
It also allowed us to generate random maps, consisting of differently spaced trees with various branch diameters.
We implemented the agent policy and the value function as multilayer perceptrons.
The developed cost function enabled successful training, giving the agent the best flight efficiency of 91\%. 

When transferring the system to a~real hardware platform, we also used the Gazebo environment.
We prepared an imitation of trees in it in the form of tall cylinders with different radii. 
In this developed environment, we ran an agent that was able to perform effective flights in a~repeatable manner.
Finally, we implemented the developed control algorithm in an Nvidia Jetson Nano chip and conducted tests on the actual drone.
In the vast majority of tests, the vehicle successfully flew through the forest, repeatedly avoiding any impact with a~tree.

It should be emphasised that the results presented in this paper represent only a~certain stage of work related to the application of reinforcement learning algorithms to the task of drone control.
Nevertheless, they are promising and justify the continuation of the chosen research direction.
Further development of this approach primarily involves the use of more graphically advanced 3D simulations.
In addition, we also plan to use sensors of other types, in particular RGB or depth cameras, as well as event (neuromorphic) sensors.
Potential improvements also include the possibility to introduce learning algorithms for the agent during real-world missions and the implementation of the entire system on other hardware platforms (e.g. heterogeneous SoC FPGAs) to speed up calculations and reduce energy consumption.

\subsection*{Acknowledgements} The work presented in this paper was supported by the AGH University of Krakow project no. 16.16.120.773.

\renewcommand\refname{References}


%

\end{document}